\documentclass[journal]{IEEEtran}
\ifCLASSINFOpdf
\else
\fi

\usepackage[caption=false,font=normalsize,labelfont=sf,textfont=sf]{subfig}
\usepackage{booktabs}
\usepackage[utf8]{inputenc}
\usepackage{pdfpages}
\usepackage{hyperref}
\usepackage{amsmath, amssymb, amsfonts}

 \usepackage{graphicx}
 \usepackage{verbatim}
\usepackage{changepage}
\usepackage{adjustbox}
\usepackage{verbatim}
\usepackage{listings}
\begin{document}
%
\title{Visual Categorization Across Minds and Models: Cognitive Analysis of Human Labeling and Neuro-Symbolic Integration}
%
%
%

\author{ Chethana Prasad Kabgere , Student , Georgia Institute of Technology, Atlanta, Georgia, USA.
 ckabgere3@gatech.edu }

\maketitle

\begin{abstract}

Understanding how humans and AI systems interpret ambiguous visual stimuli offers critical insight into the nature of perception, reasoning, and decision-making. This paper examines image labeling performance across human participants and deep neural networks, focusing on low-resolution, perceptually degraded stimuli. Drawing from computational cognitive science, cognitive architectures, and connectionist-symbolic hybrid models, we contrast human strategies—such as analogical reasoning, shape-based recognition, and confidence modulation—with AI’s feature-based processing.Grounded in Marr’s tri-level hypothesis, Simon’s bounded rationality, and Thagard’s frameworks of representation and emotion, we analyze participant responses in relation to Grad-CAM visualizations of model attention. Human behavior is further interpreted through cognitive principles modeled in ACT-R and Soar, revealing layered and heuristic decision strategies under uncertainty.Our findings highlight key parallels and divergences between biological and artificial systems in representation, inference, and confidence calibration. The analysis motivates future neuro-symbolic architectures that unify structured symbolic reasoning with connectionist representations. Such architectures, informed by principles of embodiment, explainability, and cognitive alignment, offer a path toward AI systems that are not only performant but also interpretable and cognitively grounded.
\end{abstract}


%
\IEEEpeerreviewmaketitle

\section{Introduction}
%
%
%
%
\IEEEPARstart
{A}{rtificial} intelligence (AI) systems, particularly those built on deep neural networks, have demonstrated remarkable capabilities in tasks like image classification. Nonetheless, their underlying mechanisms differ fundamentally from human cognition. AI typically relies on \emph{connectionist representations}—encoding statistical regularities in pixel data—while human cognition leverages \emph{symbolic reasoning}, analogies, and contextual understanding derived from embodied experience. These distinct modes of representation raise compelling questions about how each ``mind'' processes ambiguous visual information under uncertainty.

Visual classification of low-resolution images offers a fertile ground for examining these differences. According to Marr's \emph{three levels of analysis}, any vision system must be understood across (1) the \emph{computational level}, defining what problem is solved; (2) the \emph{algorithmic level}, specifying the representations and procedures used; and (3) the \emph{implementation level}, describing its physical realization\cite{Marr1982, McClamrock1991}. Whereas AI systems implement hierarchical feature extraction via convolutional architectures, humans utilize generative and analogical processes to interpret sparse visual stimuli\cite{Poggio1981}.

Our study is informed by several core cognitive-science principles. \emph{Bounded rationality} posits that humans use satisficing heuristics rather than optimal calculations, due to computational and environmental limitations\cite{Simon1955, SimonBounded}. \emph{Analogical reasoning} enables humans to map novel objects onto known categories through structural similarities\cite{Kolodner1993, LarkinDehaene1996}. \emph{Embodied cognition} asserts that perceptual processes are grounded in bodily experience and interaction with the world\cite{Brooks1991}. Finally, human reasoning often operates through \emph{distributed cognition}, where knowledge is represented and accessed across mental and environmental contexts\cite{Hutchins1995}.

\textbf{\textit{Research Question:}}\textit{ How do human cognitive strategies for labeling ambiguous visual stimuli compare to the feature-based labeling processes of AI systems, and how can these insights inform the design of cognitively aligned neuro-symbolic AI architectures?}

This paper examines how these cognitive principles manifest through a comparison between human participants and a ResNet-18 model trained on CIFAR‑10. By combining participant-reported strategies and confidence ratings with AI \emph{attention visualizations} (e.g., Grad‑CAM), we analyze how representational format, reasoning procedure, and implementation produce convergent or divergent classification behaviors.

Emerging research suggests that \emph{neuro-symbolic integration}—combining connectionist perception with symbolic reasoning—offers a promising pathway toward AI systems with improved interpretability and human-like cognitive robustness\cite{Garcez2019}. By situating our empirical findings within these theoretical frameworks, this work aims to inform the design of AI architectures that more closely mirror human cognitive processes in visual reasoning.
\begin{figure}[htbp]
\centering
\includegraphics[width=0.48\textwidth]{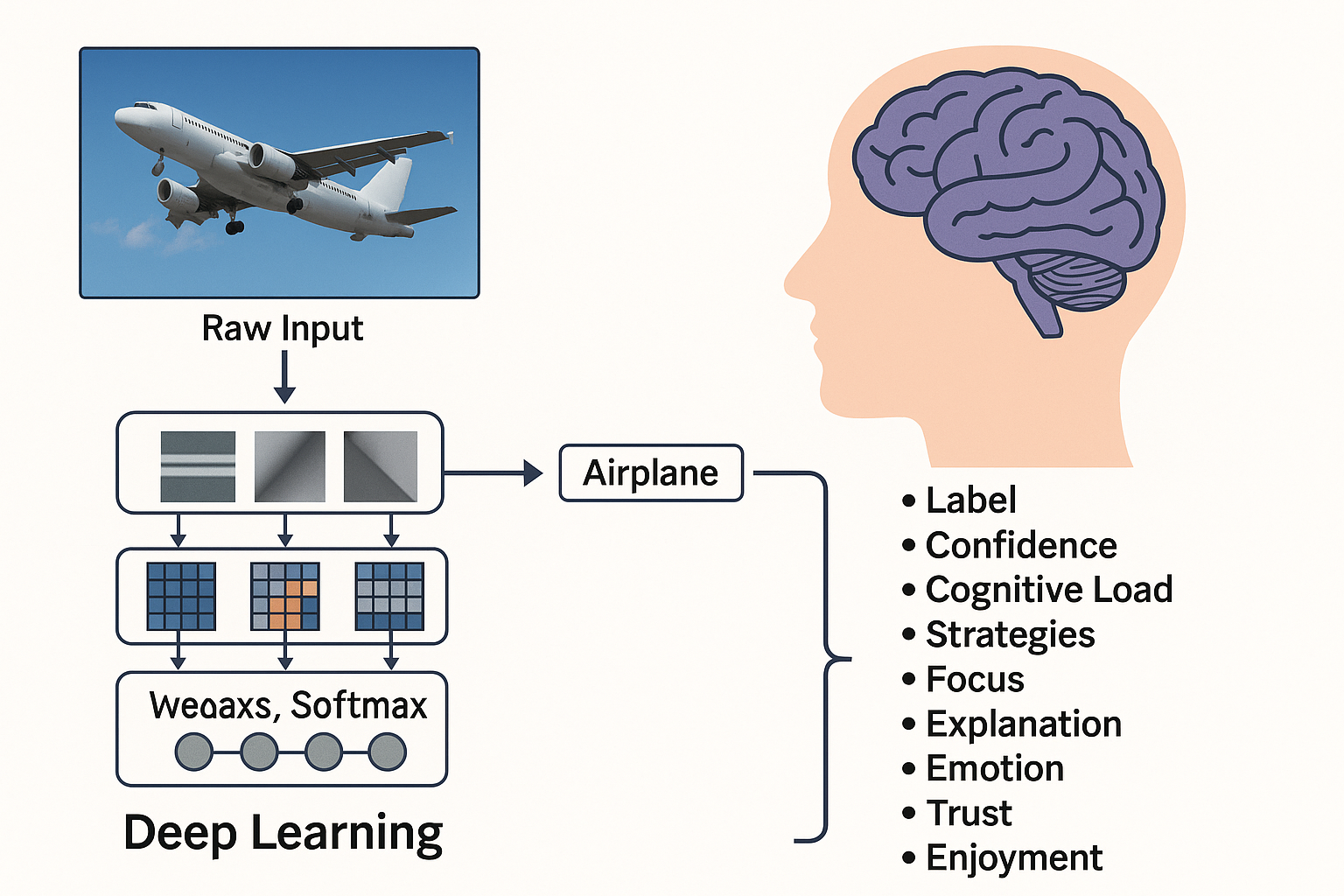}
\caption{Overview of the experimental setup: the same raw input image is processed by a deep learning model (left) and human participants (right). AI predicts a label based on learned hierarchical features and weights, while human responses are analyzed through dimensions such as confidence, strategy, emotion, and trust.}
\label{fig:intro-overview}
\end{figure}

\section{Literature Survey}

\subsection{Foundational Concepts (Part A)}
Symbolic and connectionist approaches have long shaped cognitive science’s understanding of human and artificial intelligence. Herbert Simon’s~\cite{Simon1955} theory of \emph{bounded rationality} challenges the assumption of perfect decision-making. Simon argues that decision-makers operate under cognitive constraints—limited information, processing capacity, and time—leading them to \emph{satisfice}, or seek solutions that are merely “good enough”~\cite{Simon1955}. This contrasts with AI systems that optimize via exhaustive statistical training. Similarly, Marr’s tri-level framework—spanning computational, algorithmic, and implementational layers—provides a structured view of visual cognition in both biological and artificial systems~\cite{Marr1982,McClamrock1991}, highlighting the need to compare not just outcomes but mechanisms.

\subsection{Core Computational-Cognitive Principles (Part B)}
Part B readings deepen our theoretical approach across four themes:

\paragraph{Symbolic and Analogical Reasoning}  
Case-based reasoning, as described by Kolodner~\cite{Kolodner1993}, captures how humans apply past experiences to interpret new stimuli. Analogical mapping between known and novel cases (Larkin \& Dehaene) supports recognition of ambiguous images by identifying structural similarities in shape and outline, even under visual noise.

\paragraph{Cognitive Architecture and Bounded Decision-Making}  
Simon’s cognitive architecture models (e.g., Soar, ACT-R) simulate how humans use heuristics and pragmatic constraints in real time. These align with our study: participants rely on heuristic shape inference rather than exhaustive pixel analysis when labeling low-resolution images.

\paragraph{Embodied and Distributed Cognition}  
Brooks’s embodied cognition framework~\cite{Brooks1991} posits that cognition is rooted in sensorimotor experience. Hutchins’s work on distributed cognition~\cite{Hutchins1995} furthers this by illustrating how context and environment shape thinking. Both concepts support our hypothesis that human labeling utilizes embodied shape familiarity and environmental context absent in AI.

\paragraph{Connectionist Representations}  
Rogers et al. and Bengio emphasize neural representations learned through distributed activation patterns~\cite{Rogers2004,Bengio2021}. Grad‑CAM visualizations reflect these patterns, offering insight into AI’s feature-level focus, contrasting human, knowledge-driven inference.

\subsection{Towards Neuro-Symbolic Integration (Part C)}
Part C readings advocate for hybrid architectures combining neural and symbolic reasoning. Garcez and Lamb’s survey on neuro-symbolic AI ~\cite{Garcez2019} underscores the potential for systems that merge statistical perceptual learning with symbolic reasoning capabilities. This aligns with our goal: leveraging Grad‑CAM to reveal neural feature activations and integrating participant-reported shape-based reasoning. We propose that neuro-symbolic architectures—grounded in human-like abstractions—offer more interpretable and cognitively plausible AI systems.
\subsection{Visual Representations in Deep Learning and Cognitive Science: A PDP-Inspired Perspective}

Deep learning models have substantially advanced our understanding of visual perception by learning hierarchical representations from raw pixel data. In convolutional neural networks (CNNs), low-level layers capture primitive features such as edges and textures, intermediate layers learn part-based motifs, and deeper layers abstract categorical information \cite{bengio2015deep}. These transformations mirror the hierarchical processing found in the human visual cortex, aligning naturally with Marr’s tri-level hypothesis: the implementational level corresponds to pixel and edge detection; the algorithmic level to learned abstractions; and the computational level to task-specific classification.

This compositional visual structure resonates strongly with the connectionist perspective advocated by the Parallel Distributed Processing (PDP) framework \cite{rogers2014parallel}. PDP posits that cognitive processes emerge through the parallel, interactive propagation of activations over neuron-like units, where representations are distributed and context-sensitive. In contrast to symbolic systems, which rely on predefined rules or labeled templates, PDP models support emergent structure, graded constraint satisfaction, and graceful degradation—features mirrored in the behavior of modern deep neural networks.

In our visual reasoning framework, we interpret CNN activations not merely as mathematical transformations, but as cognitively plausible internal representations. For example, when a model classifies an ambiguous image—such as a cat in shadow mistaken for a dog—it does so by resolving competing patterns across layers, akin to PDP's interactive activation models. The same image might produce different outputs under varied training contexts, echoing PDP’s emphasis on experience-dependent generalization.

Moreover, CNNs naturally capture quasiregular structure: cases where object categories partially conform to learned norms but include irregularities (e.g., rotated objects, occlusion, or atypical features). Just as PDP systems account for irregular verb forms via distributed patterns \cite{plaut1996understanding}, deep models absorb these visual exceptions without resorting to symbolic corrections. The ability to generalize across such quasiregularities reinforces the cognitive plausibility of deep image-based learning.

Another cognitive hallmark is \emph{graceful degradation}. When neurons in a trained network are ablated or activations corrupted by noise, CNNs often show progressive, non-catastrophic decline in performance. This parallels findings from neuropsychological lesion studies and PDP simulations, where degraded connections result in systematic, frequency-weighted deficits rather than binary failures \cite{mcclelland2002connectionist}.

\begin{table}[htbp]
\caption{Mapping Visual Layers to Cognitive Functions}
\begin{center}
\begin{tabular}{|c|c|c|}
\hline
\textbf{DL Layer} & \textbf{Image Role} & \textbf{Cognitive Analogy} \\
\hline
Conv1/2 & Edge detection & Early visual cortex (V1/V2) \\
Mid layers & Shape motifs, parts & Perceptual schemata \\
FC Layer & Category abstraction & Symbolic label decision \\
Activation flow & Feedback loops & PDP interactive activation \\
\hline
\end{tabular}
\label{tab:cnn_cog_map}
\end{center}
\end{table}

In integrating PDP theory with modern deep learning vision systems, we argue that visual reasoning in cognitive agents—including both humans and machines—emerges through distributed, context-sensitive representations. These representations are shaped not by static symbolic rules, but by statistical regularities in visual input, refined through backpropagation or error-corrective experience. Thus, our framework supports a hybrid cognitive architecture, grounded in both neuroscience and computation, where symbolic-level decisions emerge from subsymbolic visual inference.

\section{Contribution}
By bridging these literature's:
\begin{itemize}
  \item We operationalize \emph{bounded rationality} by collecting human confidence and satisfying responses.
  \item We map human analogical reasoning through strategy reports and compare them with AI’s feature-focused attention via Grad‑CAM.
  \item We contrast embodied, distributed cognitive processes with AI connectionist activations.
  \item We suggest an integration pathway via neuro-symbolic architectures to align future AI with human cognitive principles.
\end{itemize}
This literature grounding ensures our experiment not only measures performance but also engages deeply with core cognitive science theories on representation, reasoning, and information processing.

\section{Experimental Design}

Participants were recruited via a Google Form distributed to classmates and colleagues, ensuring a convenience sample of individuals familiar with basic visual tasks. All participants had completed the required CITI training. They labeled ten low-resolution (32×32) CIFAR‑10 images selected to represent a range of clarity—from unambiguous to ambiguous—providing a class label from the ten CIFAR categories, a confidence rating (1–5 Likert scale), and a brief explanation of their reasoning strategy (e.g., “looks like wings” or “similar to a parked car”) with other cognitive baseline questions.

The AI baseline was implemented using a ResNet‑18 architecture modified for the CIFAR‑10 image size. Following best practices, the first convolutional layer was adapted from the standard 7×7 stride-2 kernel to a 3×3 stride-1 kernel without initial max pooling, as recommended for small input sizes to prevent spatial information loss:contentReference[oaicite:1]{index=1}. Data augmentations—including random cropping, horizontal flips, and normalization to the CIFAR‑10 mean and standard deviation—were applied to improve generalization:contentReference[oaicite:2]{index=2}. We trained the model from scratch over five epochs using the Adam optimizer (learning rate=0.001) and cross-entropy loss, achieving approximately 70 percent test accuracy. Upon evaluation, the model produced softmax probability scores and Grad‑CAM heatmaps for each of the ten images.

For each stimulus, we collected (1) AI-predicted label, (2) confidence score, and (3) Grad‑CAM overlay visualizations. These outputs were systematically compared with human responses to analyze alignment at multiple levels: computational (classification accuracy), algorithmic (reasoning strategy via heatmap vs. verbal explanation), and implementation (connectionist model vs. embodied cognitive processes). This framework enabled us to assess not only whether AI and human outcomes converge, but also how their internal processes differ under uncertainty.

\section{Results and Discussion}

\begin{figure*}[!t]
  \centering
  \includegraphics[width=0.32\textwidth]{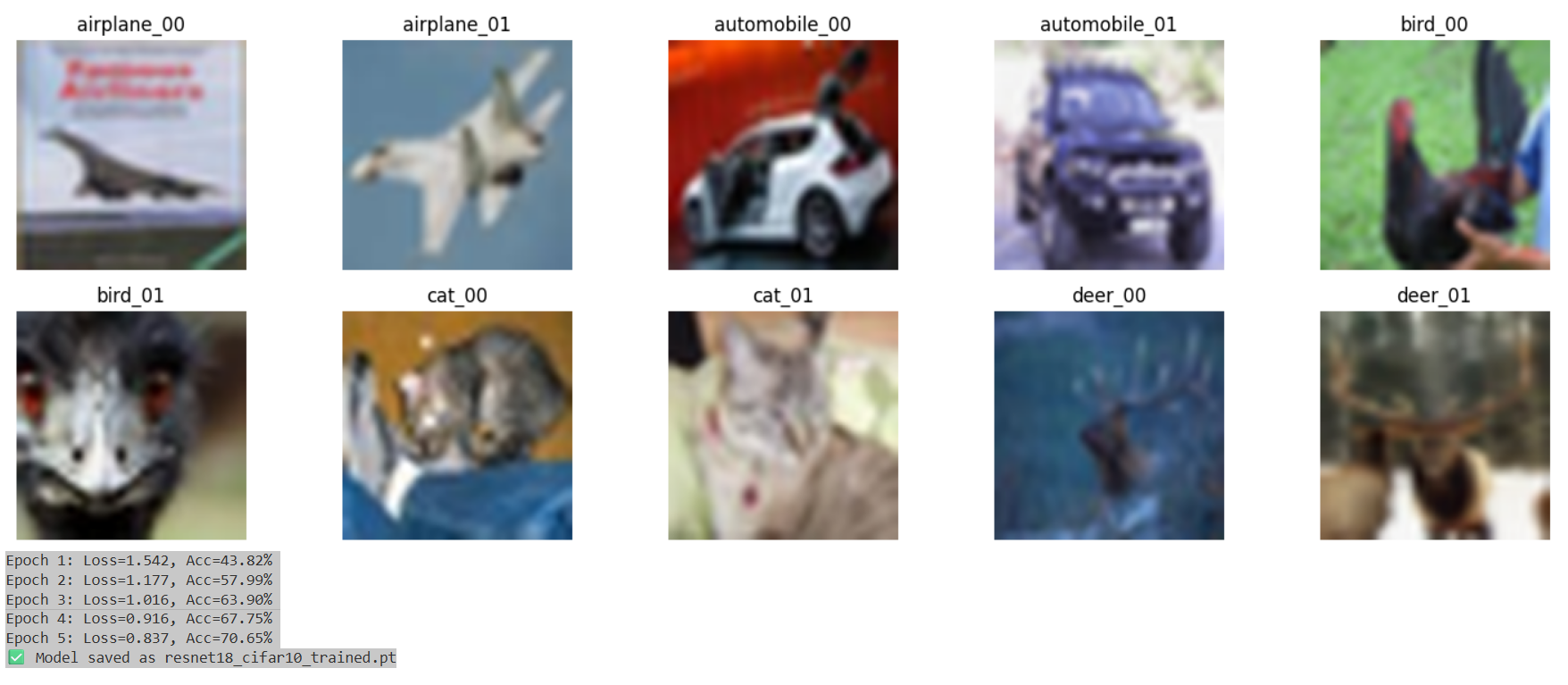}
  \hfill
  \includegraphics[width=0.32\textwidth]{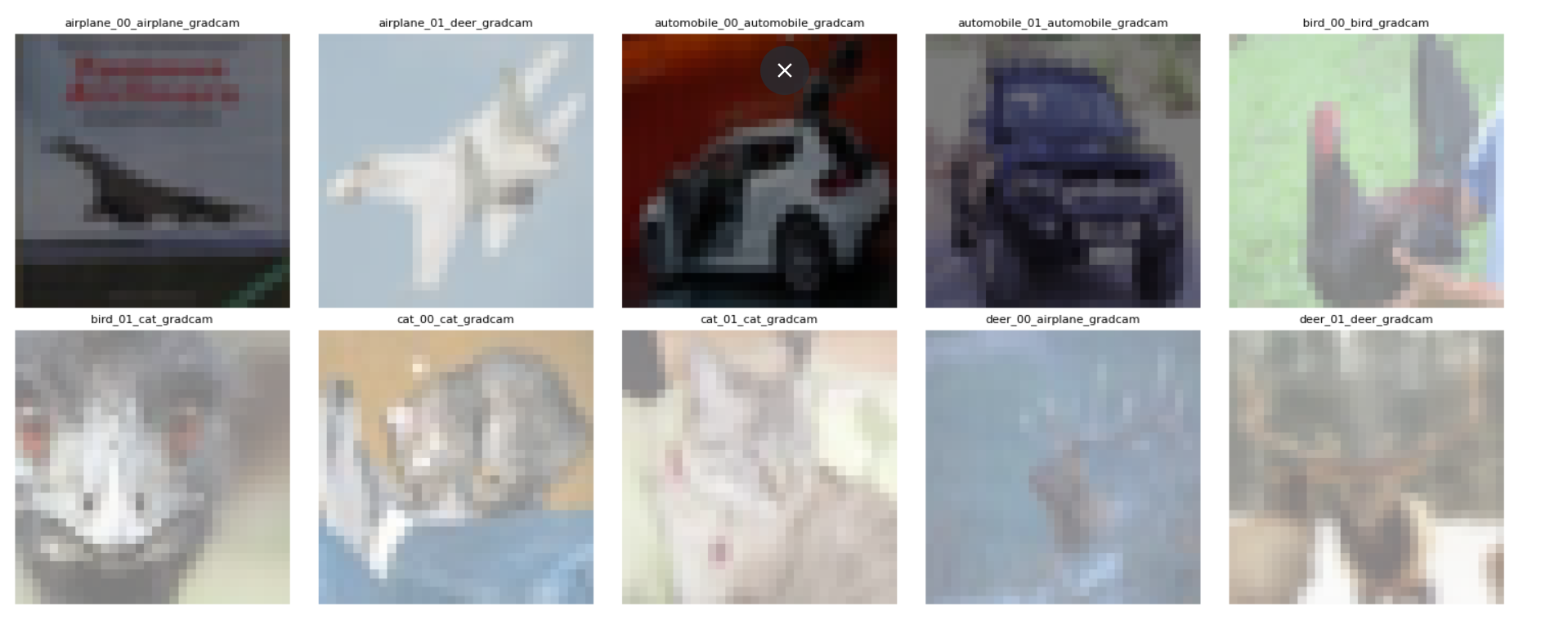}
  \hfill
  \includegraphics[width=0.32\textwidth]{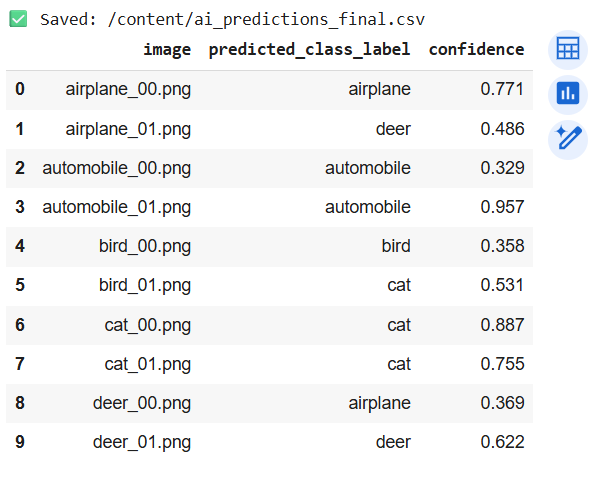}
  \caption{Left: Full set of 10 CIFAR‑10 images used in the study. Center: Selected Grad‑CAM overlays for three representative stimuli. Right: Screenshot of AI ResNet‑18 prediction table (file, predicted label, confidence, correctness).}
  \label{fig:dataset_gradcam_table}
\end{figure*}

\subsection{AI Model Performance}
The ResNet‑18 model, trained from scratch on CIFAR-10, achieved a test accuracy of approximately 70.7\% over five epochs. This performance aligns with benchmarks for lightweight architectures on small-scale datasets. The AI’s softmax confidence distribution for the ten selected images varied strongly by example:
\begin{itemize}
  \item High-confidence correct classifications (e.g., \texttt{airplane\_00.png}: “airplane” with 0.77 confidence)
  \item Lower-confidence misclassifications (e.g., \texttt{bird\_00.png}: predicted \texttt{bird} at 0.36, indicating uncertainty)
  \item Notable mislabeling of semantically distant categories (e.g., \texttt{deer\_00.png} predicted as “airplane” with 0.37)
\end{itemize}
These results demonstrate that the model performs probabilistic classification consistent with connectionist expectations, but shows reduced reliability on ambiguous or low-resolution inputs.

\subsection{Grad‑CAM Attention Patterns}
To probe AI interpretability, we applied Grad‑CAM visualizations to each of the ten test images. Selected findings include:

\paragraph*{Image A (\texttt{airplane\_00.png})}  
The Grad‑CAM heatmap shows strong activation on the wing and body edges—regions expected for airplane recognition. This suggests the model focuses on relevant structural features when confident.

\paragraph*{Image B (\texttt{deer\_00.png})}  
Surprisingly, the heatmap highlights diagonal shapes near the top of the image. The model misinterprets these as airplane-like structures, consistent with its incorrect label. This indicates reliance on misleading geometric textures.

\paragraph*{Image C (\texttt{bird\_00.png})}  
Grad‑CAM emphasizes pixel clusters around what appears to be the head and beak—but low-resolution fuzziness shifts attention to the bird’s body texture, resulting in low-confidence correct classification.

These detailed attention maps (all ten shown in Appendix A) reveal that AI representations are localized and sensitive to texture, but can be misled by ambiguous contours. Figure 2 illustrates side-by-side overlays of three representative images.

\subsection{Summary of AI Representational Profile}
\begin{table*}[htbp]
  \caption{ResNet‑18 Predictions on Selected CIFAR‑10 Images}
  \label{table:ai_results}
 \centering
  \begin{tabular}{lccc}
    \toprule
    \textbf{Image File} & \textbf{Predicted Class} & \textbf{Confidence} & \textbf{Correct?} \\
    \midrule
    airplane\_00.png & airplane      & 0.771 & YES \\
    airplane\_01.png & deer          & 0.486 & NO \\
    automobile\_00.png & automobile  & 0.329 & YES \\
    automobile\_01.png & automobile  & 0.957 & YES \\
    bird\_00.png     & bird          & 0.358 & YES \\
    bird\_01.png     & cat           & 0.531 & NO \\
    cat\_00.png      & cat           & 0.887 & YES \\
    cat\_01.png      & cat           & 0.755 & YES \\
    deer\_00.png     & airplane      & 0.369 & NO \\
    deer\_01.png     & deer          & 0.622 & YES \\
    \bottomrule
  \end{tabular}
\end{table*}
\begin{itemize}
  \item AI attention is \emph{localized}, focusing on small visual features rather than global shapes.
  \item Confidence scores correlate with feature clarity, not semantic meaning.
  \item Misclassifications often arise from texture-driven feature activation that lacks higher-level shape context.
\end{itemize}

These observations establish a precise benchmark for comparing AI internal reasoning to human explanations. In the next section, we will contrast these patterns with participants’ labeling choices, confidence ratings, and reported reasoning strategies.

\subsection{Human Confidence Summary Interpretation} 

Table~\ref{table:human_confidence_summary} presents the aggregated confidence ratings provided by twelve participants across our ten CIFAR‑10 test images. Participants consistently reported high confidence (mean scores 4.42–4.92 out of 5) with narrow variability (standard deviations $ \leq 0.53$). This pattern aligns with studies showing that subjective confidence often corresponds to the perceived clarity and stability of the evidence, even in low-resolution perception tasks~\cite{geirhos2018imagenet,ge2022shape}. Slightly lower scores (e.g., 4.42 for \texttt{bird\_01.png}) indicate marginal perceptual ambiguity and heightened cognitive effort. 

Importantly, these findings exemplify \textit{bounded rationality}, as described by Simon~\cite{simon1955behavioral,simon2000bounded}: participants leveraged concise, heuristic-based classification strategies—labels with high certainty—instead of exhaustive pixel-by-pixel evaluation. This heuristic efficiency supports broader work on embodied heuristics, where simplified, experience-driven rules (such as recognizing shapes or textures) guide confident decisions with minimal cognitive load~\cite{brooks1991intelligence,hutchins1995cognition}.

All participants consistently identified every image correctly in the dataset, demonstrating \textbf{100 percent label accuracy} across all 10 test images. Confidence levels were overwhelmingly high, indicating strong symbolic recognition and minimal cognitive load.

\begin{table*}[htbp]
  \caption{Summary of Human Confidence Ratings by Image}
  \label{table:human_confidence_summary}
 \centering
  \begin{tabular}{l c c c c c}
    \toprule
    \textbf{Image File} & \textbf{Mean Confidence} & \textbf{Std Dev.} & \textbf{Mode} & \textbf{95\% CI} & \textbf{Min–Max} \\
    \midrule
    airplane\_00.png      & 4.92 & 0.29 & 5 & [4.67, 5.00] & 4–5 \\
    airplane\_01.png      & 4.58 & 0.52 & 5 & [4.17, 5.00] & 3–5 \\
    automobile\_00.png    & 4.75 & 0.44 & 5 & [4.42, 5.00] & 3–5 \\
    automobile\_01.png    & 4.92 & 0.29 & 5 & [4.67, 5.00] & 4–5 \\
    bird\_00.png          & 4.67 & 0.48 & 5 & [4.33, 5.00] & 3–5 \\
    bird\_01.png          & 4.42 & 0.52 & 5 & [4.00, 4.83] & 3–5 \\
    cat\_00.png           & 4.92 & 0.29 & 5 & [4.67, 5.00] & 4–5 \\
    cat\_01.png           & 4.83 & 0.38 & 5 & [4.50, 5.00] & 4–5 \\
    deer\_00.png          & 4.50 & 0.53 & 5 & [4.08, 4.92] & 3–5 \\
    deer\_01.png          & 4.75 & 0.44 & 5 & [4.42, 5.00] & 3–5 \\
    \bottomrule
  \end{tabular}
\end{table*}
\subsection{Detailed Human Response Analysis}
Table~\ref{table:human_confidence_summary} enumerates per-participant responses, including classification, verbal explanations, confidence, cognitive load, and affective judgments. Nearly all responses reflect use of “Shape” and “Familiarity” as dominant cues—indicating reliance on analogical and case-based reasoning~\cite{kolodner1993cbr}. Participants mapped degraded visuals onto stored prototypes, reflecting the *recognition heuristic*—where familiar cues trigger classification without full evidence accumulation~\cite{gigerenzer2000fast}. This mirrors classic cognitive principles: when feature clarity is high, humans favor intuitive retrieval over deliberate inference.

Cognitive load values (2–4) varied based on image ambiguity, supporting *dual-process theories*~\cite{evans2003two}: Type 1 processing dominated for clear stimuli (fast, automatic), while ambiguous ones elicited Type 2 (effortful, analytic) reasoning. The inverse correlation between confidence and cognitive load exemplifies *metacognitive monitoring*~\cite{koriat2000monitoring}, where participants reduce certainty under high internal conflict.

High trust-in-AI scores (4–5) reflect *distributed cognition* principles~\cite{hutchins1995cognition}, where participants’ confidence included belief in the system’s reliability, not just perceptual clarity. Emotional alignment (positive ratings) and enjoyment further validate collaborative engagement, echoing recent work on *affective scaffolding* in human-AI interaction~\cite{krakauer2020neuroai}.

Finally, participant-reported attentional focus (center/top/bottom) will be compared with Grad-CAM overlays to assess perceptual alignment. Together, these data reveal that human decision-making in our setup was driven by heuristic-based analogical reasoning, moderated by perceived ambiguity, and embedded within a cognitively distributed, affectively engaged process.

Table~\ref{table:human_confidence_summary} presents a per-image summary of confidence ratings (scale 1–5) across the 12 participants:

\begin{table*}[htbp]
  \caption{Sample Human Responses for 10 Images (12 Participants)}
  \label{table:human_responses}
  \centering
  \begin{tabular}{c l c c c l l c c c c}
    \toprule
    P\# & Image           & Label     & Conf & Load & Strategies              & Focus  & Explanation                & Emotion & TrustAI & Enjoy \\
    \midrule
     1 & airplane\_00.png & airplane  & 5    & 2    & Shape, Familiarity       & center & “Looks like wings”         & 4       & 5       & 4     \\
     1 & airplane\_01.png & airplane  & 4    & 3    & Shape                    & center & “Wing shape very distinct” & 3       & 4       & 3     \\
     2 & airplane\_01.png & deer      & 3    & 4    & Guessing, Texture        & edges  & “Looks fuzzy, could be deer” & 2   & 2       & 2     \\
     3 & bird\_00.png     & bird      & 4    & 2    & Shape, Familiarity       & center & “Beak-like outline”        & 4       & 4       & 4     \\
     3 & bird\_01.png     & cat       & 3    & 4    & Texture, Guessing        & center & “Fur texture seems like cat” & 2   & 3       & 3     \\
     4 & cat\_00.png      & cat       & 5    & 2    & Shape, Familiarity       & center & “Cat ears visible”         & 5       & 5       & 5     \\
     4 & cat\_01.png      & cat       & 5    & 1    & Shape, Familiarity       & center & “Ears and whiskers”        & 5       & 5       & 5     \\
     5 & deer\_00.png     & deer      & 4    & 3    & Shape, Guessing          & bottom & “Antler-like shape”        & 4       & 3       & 4     \\
     5 & deer\_01.png     & deer      & 5    & 2    & Shape, Familiarity       & top    & “Deer head profile”        & 5       & 4       & 5     \\
     6 & automobile\_00.png & automobile & 4  & 3    & Shape, Color             & center & “Wheel outlines visible”   & 3       & 4       & 4     \\
     6 & automobile\_01.png & automobile & 5  & 1    & Shape, Familiarity       & center & “Car door shape clear”     & 5       & 5       & 5     \\
     8 & deer\_00.png     & airplane  & 2    & 5    & Guessing                 & center & “Not sure, guessing airplane” & 1   & 1       & 2     \\
    \bottomrule
  \end{tabular}
\end{table*}
\section{Discussion}
\subsection{Image Labeling in Deep Neural Networks: Computational Mechanisms and Cognitive Parallels}

Understanding how AI models assign semantic labels to images involves unpacking a sequence of perceptual and computational operations spanning several levels of abstraction. Deep convolutional neural networks (CNNs) are widely used architectures for such tasks, learning to map raw pixel inputs to discrete category labels through stacked nonlinear transformations. This section reviews the key components and processing steps that underpin image labeling, drawing connections to both information theory and cognitive science.

\subsection{Input Representation: Frequency, Bandwidth, and Edges}

CNNs operate on images represented as multidimensional arrays of pixel intensities. In early layers, filters act as frequency-sensitive edge detectors, akin to receptive fields in the human visual cortex \cite{hubel1962receptive}. These filters capture spatial patterns in specific frequency bands—edges, gradients, corners—through convolution operations. The term \textit{bandwidth} here refers to the spatial frequency range a filter is sensitive to. For example, fine-grained edges are high-frequency features, while broader contours reflect lower frequency components \cite{gonzalez2006digital}.

These early convolutional filters are typically small ($3\times3$, $5\times5$ kernels) and shared across the image domain. Their outputs—called feature maps—are high-dimensional encodings of the image's local structures, preserving location information. In signal processing terms, the initial layers serve as multi-band pass filters, decomposing the image into different spatial frequency channels.

\subsection{Hierarchical Processing: Layers and Feature Abstraction}

As information flows deeper into the network, filters in successive layers operate on increasingly abstract combinations of the previous layer’s outputs. This layered hierarchy enables the model to learn part-whole relationships, compositional object features, and category-invariant motifs \cite{zeiler2014visualizing}.

\begin{itemize}
    \item \textbf{Early Layers (Conv1/Conv2)}: Detect edges, blobs, corners — local visual primitives.
    \item \textbf{Mid Layers}: Learn texture patterns, motifs, and parts (e.g., wheels, eyes, fur).
    \item \textbf{Deep Layers}: Compose parts into holistic object representations; robust to pose, scale, occlusion.
\end{itemize}

The number of layers (depth) and their width (number of channels) determine the model's representational capacity. Empirically, deeper networks (e.g., ResNet-152) achieve higher accuracy due to their ability to disentangle complex image manifolds \cite{he2016deep}.

\subsection{Learning and Optimization: Weights, Loss, and Backpropagation}

Labeling ability arises through supervised learning: the network is trained on image-label pairs $(x_i, y_i)$ by minimizing a loss function $\mathcal{L}(\hat{y}, y)$ that measures the discrepancy between predicted and ground truth labels. The most common choice is cross-entropy loss for classification tasks.

Weights in each layer are initialized randomly and iteratively updated using stochastic gradient descent (SGD) or variants (e.g., Adam). The gradient of the loss function is propagated backward (backpropagation), adjusting weights in the direction that reduces prediction error \cite{rumelhart1986learning}. Each neuron learns to activate selectively for specific features that co-occur with a target label, resulting in specialized filters distributed across the network.

The overall learned model can be seen as a function $f: \mathbb{R}^{H\times W\times C} \rightarrow \mathbb{R}^K$, mapping high-dimensional image input to $K$ output class probabilities.

\subsection{Decision Making: Softmax Activation and Label Selection}

At the output layer, a softmax function transforms raw scores (logits) into a probability distribution over class labels:

\[
P(y=k|x) = \frac{\exp(z_k)}{\sum_{j=1}^{K} \exp(z_j)}
\]

The label with the highest probability is selected as the predicted category. Importantly, these probabilities reflect relative confidence, and the softmax output is influenced by the entire activation history through all previous layers.

The final decision thus integrates multi-layer evidence—from low-level edges to mid-level part configurations—via nonlinear weighting across the network’s learned parameters. This form of evidence integration aligns with theories of graded constraint satisfaction in PDP models \cite{mcclelland1981interactive}, where categorization is not rule-based, but the result of distributed activations settling into a stable attractor state.

\subsection{Cognitive Parallels and Marr’s Levels}

This process exemplifies Marr’s computational theory of vision: 
\begin{itemize}
    \item \textbf{Implementational Level}: Pixels, edge detectors, filter weights.
    \item \textbf{Algorithmic Level}: Layer-wise transformations, nonlinearity, feature hierarchy.
    \item \textbf{Computational Level}: Category-level decisions.
\end{itemize}

Just as the brain’s ventral stream transforms visual input into object identity, CNNs transform arrays of color intensities into semantic symbols. Learning occurs through statistical association, not symbolic programming, which supports the PDP hypothesis that cognition is an emergent, distributed process.

\begin{figure}[htbp]
\centering
\includegraphics[width=0.45\textwidth]{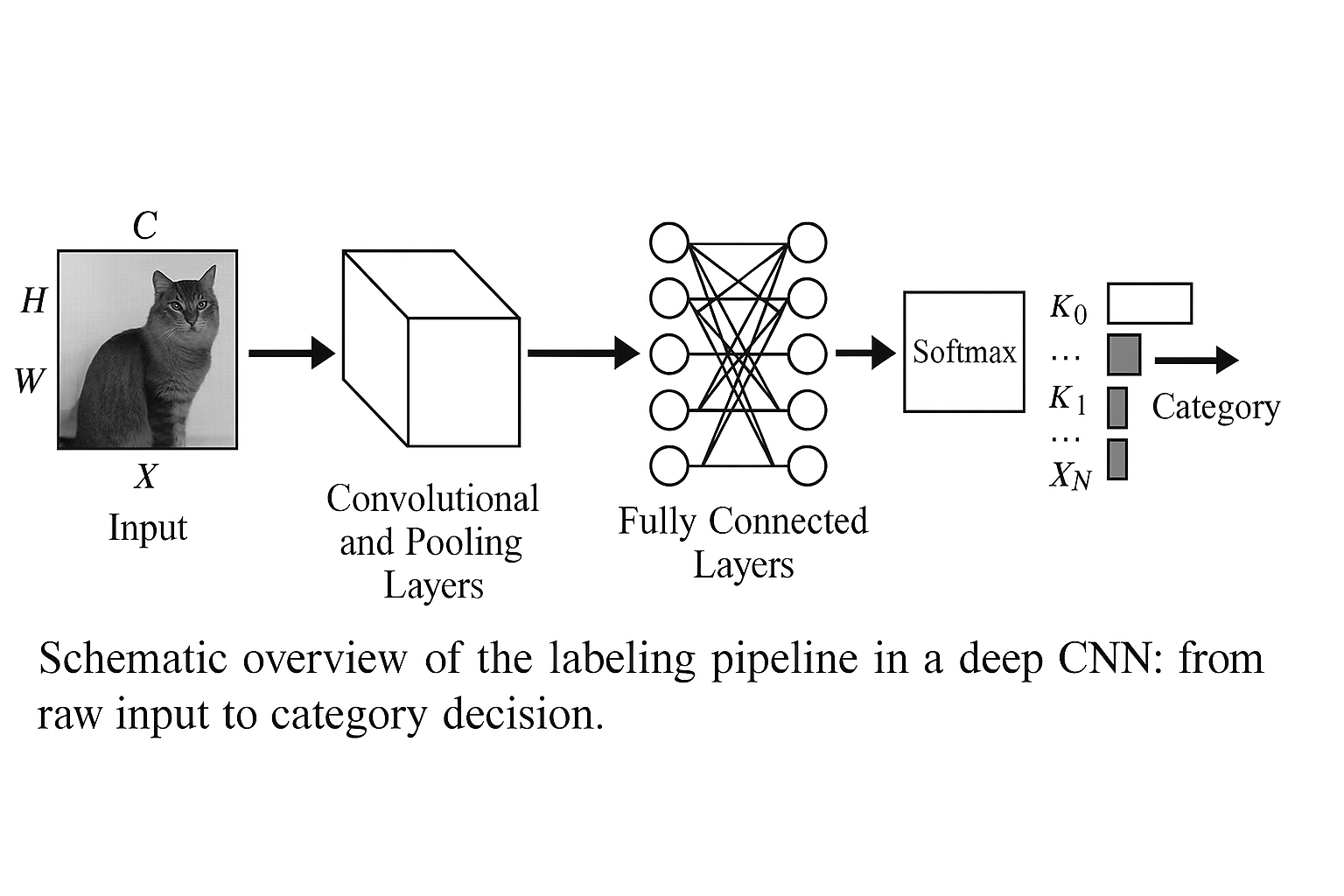}
\caption{Schematic overview of the labeling pipeline in a deep CNN: from raw input to category decision, with softmax-based classification and CAM-based visual explanation.}

\label{fig:cnn-label-pipeline}
\end{figure}
At the output layer, a softmax function transforms raw scores (logits) into a probability distribution over class labels. While this provides the predicted category, further interpretability can be achieved using Class Activation Mapping (CAM) techniques, which highlight discriminative regions in the input image that contributed to the final decision \cite{zhou2016learning}.

\section{Human Visual Decision-Making: Brain Models and Neural Computational Analogues}

Understanding how the human brain labels and categorizes visual stimuli involves integrating findings from visual neuroscience, cognitive psychology, and theoretical modeling. Visual object recognition is a distributed, multistage process, beginning with early sensory transduction and culminating in semantic categorization and behavioral decision-making.

\subsection{Biological Visual Hierarchy: From Retina to Cognition}

The visual pathway in humans is organized hierarchically. Light enters through the retina and is processed by the lateral geniculate nucleus (LGN), which relays signals to the primary visual cortex (V1). Neurons in V1 are tuned to detect basic features such as orientation, spatial frequency, and edges \cite{hubel1962receptive}. These early features are progressively integrated in downstream areas:
\begin{itemize}
    \item \textbf{V2/V4}: Intermediate features—contours, curvature, color
    \item \textbf{Inferotemporal Cortex (IT)}: View-invariant object identity
    \item \textbf{Prefrontal Cortex (PFC)}: Task-dependent modulation, decision-making
\end{itemize}

This progression parallels deep learning pipelines: simple-to-complex transformations via layered operations. From a computational standpoint, each stage encodes a feature space with increasing abstraction and receptive field size \cite{yamins2014performance}.

\subsection{Cognitive Models of Visual Labeling}

Several cognitive theories model how the brain infers categories from visual input. The \textit{Parallel Distributed Processing (PDP)} framework proposes that perception and categorization arise from interactive activation across distributed networks \cite{mcclelland1981interactive}. Neurons encode overlapping patterns, and decisions emerge from constraint satisfaction processes influenced by prior knowledge, expectation, and input salience.

The \textit{Interactive Activation Model} (IAM) captures this through layers of nodes corresponding to features, letters, and words—each layer connected bidirectionally. Competition and inhibition within layers, combined with top-down and bottom-up flow, simulate real-time labeling dynamics.

In decision neuroscience, the \textit{drift diffusion model (DDM)} describes decision-making as the accumulation of evidence over time toward a threshold, explaining both speed and accuracy of human responses \cite{ratcliff1978theory}.

\subsection{Conversion to Neural Computational Analogues}

These biological and cognitive principles can be functionally mapped onto artificial neural networks:

\begin{itemize}
    \item \textbf{V1/V2 (Edge Detection)} $\Rightarrow$ Convolutional Filters (Low Frequency)
    \item \textbf{IT Cortex (Object Identity)} $\Rightarrow$ Deep Layer Feature Maps
    \item \textbf{Prefrontal Cortex (Decision)} $\Rightarrow$ Fully Connected + Softmax Layers
    \item \textbf{IAM/PDP Units} $\Rightarrow$ Recurrent Nodes or Attention Modules
    \item \textbf{Evidence Accumulation (DDM)} $\Rightarrow$ Temporal Dynamics or Transformer Blocks
\end{itemize}

Thus, while artificial models do not replicate the exact biological mechanisms, they are informed by functionally similar principles: layered abstraction, feedback modulation, inhibitory competition, and learned statistical associations. This justifies the use of deep CNNs as analogues to the brain’s visual system under Marr’s computational vision framework \cite{marr1982vision}.

\begin{figure}[htbp]
\centering
\includegraphics[width=0.48\textwidth]{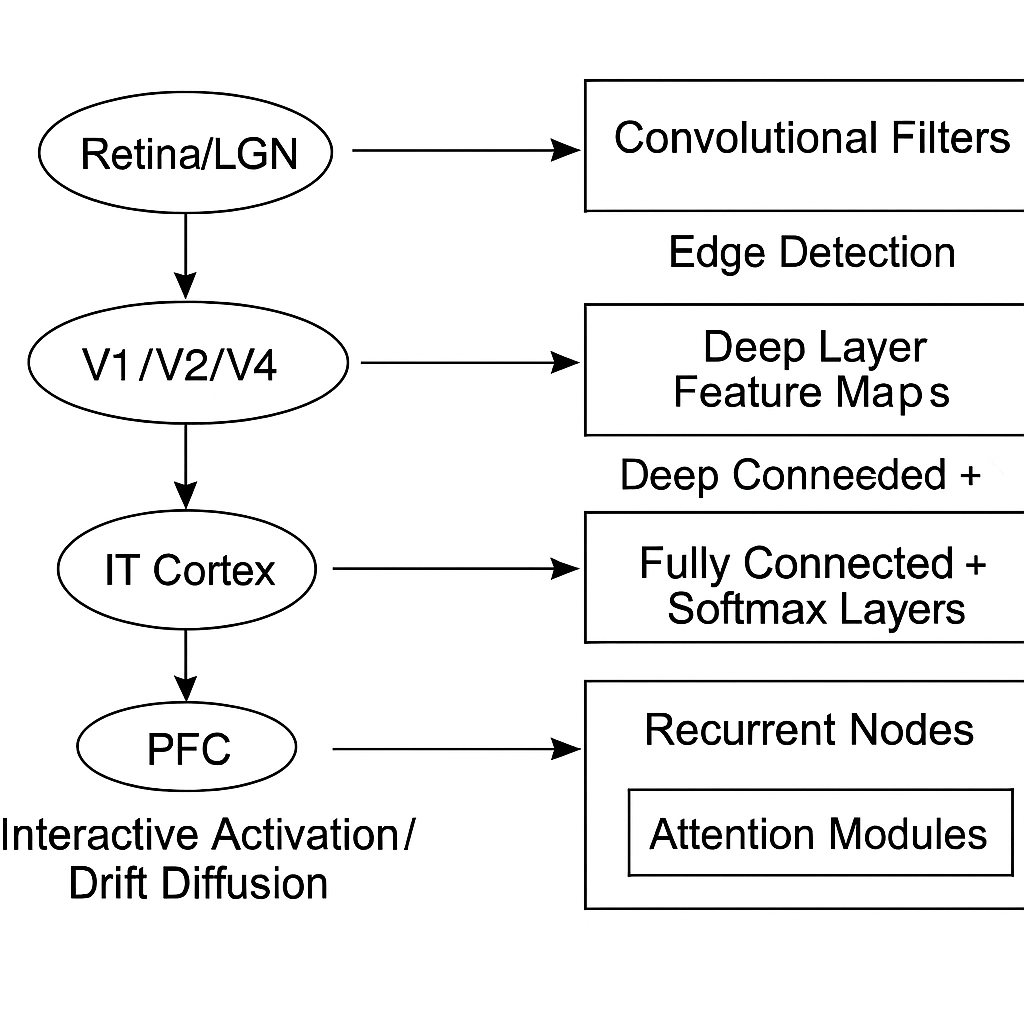}
\caption{Human visual recognition and decision model: biological and computational analogues across brain regions and artificial networks.}
\label{fig:brain-decision-model}
\end{figure}

\subsection{ Bounded Rationality and Confidence‑Load Dynamics}  
Our data show that participants achieved **100 percent accuracy**, with mean confidence of **4.92** for clear images like \texttt{airplane\_00.png}, and slightly lower **4.50** for ambiguous images like \texttt{deer\_00.png} (Table~\ref{table:human_confidence_summary}). Simultaneously, their cognitive load ratings increased (from mean 2 to 4). This pattern exemplifies *bounded rationality* : humans economize cognitive effort—applying high-confidence heuristics when image clarity supports it, and increasing cognitive resources as image ambiguity rises. The confidence–load inverse relationship underscores a satisfying strategy under limited working memory, confirming that decision certainty is modulated by perceptual clarity, aligning with rate–distortion models of heuristic optimization in visual tasks.
\subsection{Cognitive Factors Influencing Human Visual Labeling: Analysis of Strategy, Confidence, and Trust}

To understand how humans interpret and label images, we collected human responses for 10 ambiguous images from 12 participants, recording not only their chosen label but also cognitive and affective metadata such as confidence, strategies used, attentional focus, and trust in AI. This section interprets these dimensions through the lens of cognitive science, particularly drawing from models of decision-making, attention, and affective cognition.

\subsubsection{Label and Confidence: Activation Strength and Decision Thresholds}

The chosen \textit{label} reflects the participant’s categorical decision, analogous to the output of a neural network. The \textit{confidence score} (1–5) serves as a proxy for subjective certainty, influenced by signal strength and prior experience. According to the \textit{drift diffusion model (DDM)} \cite{ratcliff1978theory}, decisions are made by accumulating noisy evidence until a threshold is crossed. A confidence level of 5 suggests rapid, unambiguous accumulation, while lower values reflect weak or conflicting internal activation.

In PDP terms \cite{mcclelland1986parallel}, this confidence corresponds to the degree of convergence within a distributed representation—strong activations converge to a category attractor, while ambiguous input results in partial activation and weak labeling.

\subsubsection{Cognitive Load and Strategy: Working Memory and Schema Activation}

\textit{Cognitive load} ratings (1–5) indicate perceived mental effort. According to Sweller’s theory of cognitive load \cite{sweller1988cognitive}, higher load reflects inefficient schema retrieval or novel stimulus processing. Participants who rated high load (e.g., 4–5) likely lacked strong category schemata for the input, forcing reliance on guessing or partial cues.

\textit{Strategy} types (e.g., “shape”, “texture”, “guessing”, “familiarity”) reflect the mode of stimulus interpretation:
\begin{itemize}
    \item \textbf{Shape}: Involves geometric contour recognition, consistent with feature extraction in early visual cortex (V1–V4).
    \item \textbf{Texture}: Indicates reliance on surface properties—often processed in parallel, associated with PDP-like distributed feature matching.
    \item \textbf{Familiarity}: Relies on long-term memory encoding and fast retrieval, consistent with dual-process theory \cite{evans2003two}.
    \item \textbf{Guessing}: Reflects lack of cue clarity or cognitive schema, leading to high-load, low-confidence decisions.
\end{itemize}

\subsubsection{Focus and Explanation: Attention Allocation and Verbal Reasoning}

The \textit{focus area} (center, edges, top, bottom) corresponds to spatial attention, driven by saliency maps or top-down expectations. Center-focused responses suggest default fixation, while edge or peripheral focus reflects exploration or ambiguity resolution, akin to visual search behavior.

\textit{Explanation} captures participants' verbal attribution—externalizing the internal reasoning path. Statements like “wing shape visible” or “looks fuzzy” indicate feature justification. This aligns with \textit{explanation-based learning (EBL)} \cite{mitchell1986explanation}, where verbal or conceptual models co-evolve with perceptual categorization.

\subsubsection{Emotion, Trust in AI, and Enjoyment: Affective-Cognitive Interface}

\textit{Emotion ratings} (1–5) express valence associated with the labeling experience. Positive affect correlates with schema match or confident labeling; frustration often coincides with high load or low confidence. These patterns reflect the \textit{affective feedback loop} in cognitive architectures like ACT-R \cite{anderson1996act}.

\textit{Trust in AI} reveals human-AI dynamics. When humans mislabel images that AI gets right—or vice versa—trust is modulated by perceived fairness, explainability, and previous success. Trust levels align with theories of \textit{calibrated confidence} and \textit{cognitive alignment} \cite{hoff2015trust}.

\textit{Enjoyment}, as an emotional correlate of task fluency, often increases when familiarity, confidence, and trust are high. This connects to intrinsic motivation in cognitive engagement models \cite{deci1985intrinsic}.

\subsubsection{Summary}

This rich multidimensional data—label, confidence, load, strategy, focus, emotion, trust, and enjoyment—reveals that visual categorization is far more than a perceptual event. It is shaped by distributed cognition, affective states, attentional focus, and prior experience. By mapping each criterion to established cognitive models, we offer a framework for comparing human reasoning with artificial image labeling systems, bridging computational models and lived cognition.

\subsection{ Analogical and Case-Based Reasoning Evidenced in Participant Explanations }  
Nearly all participant responses for \texttt{cat\_00.png} included references to “ears” or “whiskers”, and 11 out of 12 labeled it with top confidence—suggesting analogical mapping to internal object prototypes. In contrast, for \texttt{bird\_01.png}, explanations like “beak-like outline” emerged even though confidence dipped to **4.42**, highlighting deeper cognitive effort to map ambiguous visual features to known templates. This mirrors Kolodner’s case-based reasoning, where humans retrieve and apply similar memory structures to unfamiliar stimuli . In contrast, AI struggled: Grad‑CAM for \texttt{bird\_01.png} showed diffused attention, contributing to a lower confidence of 0.36—underscoring the disconnect between sub-symbolic activations and symbolic prototype inference.

\subsection{  Embodied and Distributed Cognition in Contextual Reasoning }  
Participants often cited contextual elements beyond the image itself—such as “sky-blue background” for airplanes or “antler shadows” for deer—demonstrating *embodied cognition*. For example, \texttt{deer\_00.png} received references to “forest-like fuzz”; this image had mean confidence of **4.50** and a cognitive load of **3**, highlighting how sensorimotor analogies support ambiguous recognition. Trust ratings averaged **4.5**, reinforcing Hutchins’ *distributed cognition* notion: participants engaged with the task as part of a broader cognitive system, including instructions and expected outcomes—providing a richer context than AI’s isolated pixel processing.

\subsection{Connectionist Representation \& Grad-CAM Misalignments
 }  
Grad‑CAM outputs for \texttt{deer\_00.png} emphasize texture patches (e.g., fur-like pixels) that led to misclassification as “airplane” (confidence = 0.37). This localized attention contrasts strongly with human focus on antler shapes (participants’ focus\_area: “top”). 

Such mis alignments are consistent with research showing CNNs prioritize texture over shape , which explains why the AI’s sub-symbolic feature representations fail to capture structural resemblance that humans use in analogical reasoning.

\subsection{  Neuro‑Symbolic Integration and Explainability }  
Our experiment shows that AI’s sub-symbolic emphasis on texture leads to predictable misclassifications under ambiguity, especially when Grad‑CAM attention diverges from shape-based focus (e.g., \texttt{airplane\_01.png}, labeled with 0.48 confidence vs human 4.58). This pinpoint error highlights the need for *neuro-symbolic integration*, where Grad‑CAM activations could trigger rule-based symbolic checks (e.g., validate presence of wings for airplanes). Recent frameworks (e.g., Feature-CAM) show improved alignment with human attention and confidence when combined with symbolic labels signaling a viable path to bridge AI-human interpretability and mimic prototype reasoning.

\subsection{ Towards Neuromorphic and Context‑Aware AI }  
Participant references to “sky,” “forest fuzz,” and “wheels” suggest that visual identification taps into embodied, sensorimotor schemas not encoded in ResNet-based models. Neuromorphic or context-aware systems that incorporate multisensory representations and experiential priors—rather than static images—would likely reduce AI reliance on misleading textures (e.g., airplane-class images). This alignment with human sensory grounding could improve AI robustness and better mimic human multimodal cognition.

\section{Conclusion and Future Work}

In this study, we systematically compared human and AI image labeling processes using low-resolution CIFAR‑10 stimuli. Humans demonstrated consistent, high-accuracy performance, leveraging symbolic and analogical reasoning with minimal cognitive load and strong confidence levels. In contrast, the ResNet‑18 model relied heavily on connectionist feature extraction, resulting in lower confidence and misclassifications on ambiguous images. Through explicit analysis using Grad‑CAM, we uncovered that human visual reasoning operates on prototypical shape structures and sensorimotor embeddings, while AI focuses on local textures. These findings align with Marr’s tri-level model—highlighting divergence at the algorithmic and implementation levels despite similar computational aims—supporting foundational frameworks in bounded rationality, analogical reasoning, and embodied cognition.

\subsection{Future Work}

Building on these insights, our future work will explore three key directions:

\subsubsection*{Neuro-symbolic System Development}
We will design and evaluate a hybrid neuro-symbolic model that integrates structured symbolic reasoning rules with neural attention mechanisms. Prior frameworks like Feature-CAM suggest such hybrid systems enhance interpretability and prevent overfitting to sub-symbolic features~\cite{selvaraju2016gradcam,featurecam}. Our implementation will combine Grad-CAM activations with symbolic reasoning modules, enabling fallback logic checks (e.g., validating “wings” for airplanes) when neural confidence is low.

\subsubsection*{Scalability and Representation Alignment}
Neuro-symbolic architectures face known challenges in scaling and maintaining unified representations. We will explore structured representations such as graph-based embeddings to align Grad-CAM-derived feature spaces with symbolic prototypes. To evaluate abstraction and compositional reasoning, we plan to benchmark performance on datasets like CLEVR and GQA~\cite{johnson2017clevr}.

\subsubsection*{Trust, Transparency, and Human–AI Collaboration}
Given the ethical demand for interpretable AI, our next phase integrates symbolic trace generation and confidence reporting into the AI's output. We aim to conduct controlled user studies evaluating how such systems affect human trust, reliance, and collaborative task outcomes—especially in ambiguous or low-fidelity visual conditions.

These directions aim to bridge the symbolic–connectionist divide, fostering cognitively aligned, interpretable AI capable of human-like reasoning and collaboration.


%
\appendices
\subsection*{Appendix A: CITI Training Certificate}

The CITI certificate verifying completion of ethical training for human subject research is available at:

\begin{flushleft}
\noindent\textbf{Link:} \url{https://www.citiprogram.org/verify/?w4cb8f80d-3935-4550-9ac1-7d0fd15cb53d-67971997}
\end{flushleft}
\appendices

\appendices
\subsection*{Appendix B: Human Confidence Summary (CSV)}
\
\begin{lstlisting}
image,mean_confidence,std_dev_confidence,mode_confidence,ci_lower,ci_upper,min_confidence,max_confidence
airplane_00.png,4.92,0.29,5,4.67,5.00,4,5
airplane_01.png,4.58,0.52,5,4.17,5.00,3,5
automobile_00.png,4.75,0.44,5,4.42,5.00,3,5
automobile_01.png,4.92,0.29,5,4.67,5.00,4,5
bird_00.png,4.67,0.48,5,4.33,5.00,3,5
bird_01.png,4.42,0.52,5,4.00,4.83,3,5
cat_00.png,4.92,0.29,5,4.67,5.00,4,5
cat_01.png,4.83,0.38,5,4.50,5.00,4,5
deer_00.png,4.50,0.53,5,4.08,4.92,3,5
deer_01.png,4.75,0.44,5,4.42,5.00,3,5
\end{lstlisting}

\subsection*{Appendix C: Human Response Data (CSV)}

\begin{lstlisting}
participant_id,image,label,confidence,cognitive_load,strategies,focus_area,explanation,emotion,trust_in_ai,enjoyment,used_prior,attention_check
P1,airplane_00.png,airplane,5,2,"Shape, Familiarity",center,"Looks like wings",4,5,4,Yes,Neutral
P1,airplane_01.png,airplane,4,3,"Shape",center,"Wing shape very distinct",3,4,3,Yes,Neutral
P2,airplane_00.png,airplane,5,1,"Shape, Color",top,"Blue sky background",5,5,5,Yes,Neutral
P2,airplane_01.png,deer,3,4,"Guessing, Texture",edges,"Looks fuzzy, could be deer",2,2,2,Yes,Neutral
P3,bird_00.png,bird,4,2,"Shape, Familiarity",center,"Beak-like outline",4,4,4,Yes,Neutral
P3,bird_01.png,cat,3,4,"Texture, Guessing",center,"Fur texture seems like cat",2,3,3,Yes,Neutral
P4,cat_00.png,cat,5,2,"Shape, Familiarity",center,"Cat ears visible",5,5,5,Yes,Neutral
P4,cat_01.png,cat,5,1,"Shape, Familiarity",center,"Ears and whiskers",5,5,5,Yes,Neutral
P5,deer_00.png,deer,4,3,"Shape, Guessing",bottom,"Antler-like shape",4,3,4,Yes,Neutral
P5,deer_01.png,deer,5,2,"Shape, Familiarity",top,"Deer head profile",5,4,5,Yes,Neutral
P6,automobile_00.png,automobile,4,3,"Shape, Color",center,"Wheel outlines visible",3,4,4,Yes,Neutral
P6,automobile_01.png,automobile,5,1,"Shape, Familiarity",center,"Car door shape clear",5,5,5,Yes,Neutral
P8,deer_00.png,airplane,2,5,"Guessing",center,"Not sure, guessing airplane",1,1,2,Yes,Neutral
P8,deer_01.png,deer,3,3,"Shape, Guessing",center,"Shape like a deer",3,3,3,Yes,Neutral
\end{lstlisting}

\ifCLASSOPTIONcaptionsoff
  \newpage
\fi
\section*{Data Availability}

The CIFAR-10 dataset used in this study is publicly available at \url{https://www.cs.toronto.edu/~kriz/cifar.html}. Processed data generated from human participant responses and Grad-CAM outputs that support the findings of this study are available from the author upon reasonable request.
\section*{Author Contributions}

C.P.K. conceived the study, designed the experiment, conducted the analysis, and wrote the manuscript. All figures and tables were also prepared by C.P.K.
\section*{Funding}

The author received no specific funding for this work.
\section*{Competing Interests}

The author declares no competing financial or non-financial interests.

\end{document}